\def\year{2021}\relax
\documentclass[letterpaper]{article} 
\usepackage{aaai21}  
\usepackage{times}  
\usepackage{helvet} 
\usepackage{courier}  
\usepackage[hyphens]{url}  
\usepackage{graphicx} 
\usepackage{graphicx}  
\usepackage{natbib}  
\usepackage{caption} 
\usepackage{comment}
\usepackage{amsmath}
\usepackage{bbm}
\usepackage{setspace}
\usepackage{multirow}
\usepackage{booktabs}
\usepackage{amsfonts}
\usepackage{siunitx}
\usepackage{makecell}
\usepackage[switch]{lineno}

\newtheorem{definition}{Definition}

\title{Meta-Transfer Learning for Low-Resource Abstractive Summarization}
\author {

        Yi-Syuan Chen,
        Hong-Han Shuai \\
}
\affiliations {
    National Chiao Tung University, Taiwan \\
    yschen.eed09g@nctu.edu.tw, hhshuai@nctu.edu.tw
}

\begin{document}
\maketitle

\begin{abstract}
Neural abstractive summarization has been studied in many pieces of literature and achieves great success with the aid of large corpora. However, when encountering novel tasks, one may not always benefit from transfer learning due to the domain shifting problem, and overfitting could happen without adequate labeled examples. Furthermore, the annotations of abstractive summarization are costly, which often demand domain knowledge to ensure the ground-truth quality. Thus, there are growing appeals for \textit{Low-Resource Abstractive Summarization}, which aims to leverage past experience to improve the performance with limited labeled examples of target corpus. In this paper, we propose to utilize two knowledge-rich sources to tackle this problem, which are large pre-trained models and diverse existing corpora. The former can provide the primary ability to tackle summarization tasks; the latter can help discover common syntactic or semantic information to improve the generalization ability. We conduct extensive experiments on various summarization corpora with different writing styles and forms. The results demonstrate that our approach achieves the state-of-the-art on 6 corpora in low-resource scenarios, with only 0.7\% of trainable parameters compared to previous work. 
\end{abstract}

\section{Introduction}
The goal of neural abstractive summarization is to comprehend articles and generate summaries that faithfully convey the core idea. Different from extractive methods, which summarize articles by selecting salient sentences from the original text, abstractive methods~\cite{Song_Shuai_Yeh_Wu_Ku_Peng_2020,Chen_Chen_Shuai_Peng_2020,gehrmann-etal-2018-bottom,see-etal-2017-get,rush-etal-2015-neural} are more challenging and flexible due to the ability to generate novel words. However, the success of these methods often relies on a large number of training data with ground-truth, while generating the ground-truth of summarization is highly complicated and often requires professionalists with domain knowledge. Moreover, different kinds of articles are with various writing styles or forms, e.g., news, social media posts, and scientific papers. Therefore, the \textit{Low-Resource Abstractive Summarization} has emerged as an important problem in recent years, which aims to leverage related sources to improve the performance of abstractive summarization with limited target labeled examples.

Specifically, to tackle data scarcity problems, a recent line of research~\cite{ijcai2020-553,pmlr-v119-zhang20ae,rothe-etal-2020-leveraging,lewis-etal-2020-bart,liu-lapata-2019-text} leverages the large pre-trained language models~\cite{NEURIPS2020_1457c0d6,devlin-etal-2019-bert}, which are trained in a self-supervised way based on unlabeled corpora. Since many natural language processing tasks share common knowledge in syntax, semantics, or structures, the pre-trained model has attained great success in the downstream summarization tasks. Another promising research handling low-resource applications is meta learning. For example, the recently proposed Model-Agnositc Meta Learning (MAML)~\cite{pmlr-v70-finn17a} performs well in a variety of NLP tasks, including machine translation \cite{Li_Wang_Yu_2020,gu-etal-2018-meta}, dialog system \cite{qian-yu-2019-domain,madotto-etal-2019-personalizing}, relational classification \cite{obamuyide-vlachos-2019-model}, semantic parsing \cite{guo-etal-2019-coupling}, emotion learning \cite{zhao-ma-2019-text}, and natural language understanding \cite{dou-etal-2019-investigating}. Under the assumption that similar tasks possess common knowledge, MAML condenses shared information of source tasks into the form of weight initialization. The learned initialization can then be used to learn novel tasks faster and better. 

Based on these observations, we propose to integrate the self-supervised language model and meta learning for low-resource abstractive summarization. However, three challenges arise when leveraging the large pre-trained language models and meta learning jointly. First, most state-of-the-art summarization frameworks~\cite{pmlr-v119-zhang20ae,rothe-etal-2020-leveraging,liu-lapata-2019-text} exploit Transformer based architecture, which possesses a large number of trainable parameters. However, the training data size of the tasks in MAML is often set to be small, which may easily cause overfitting for a large model \cite{pmlr-v97-zintgraf19a}. Second, MAML could suffer from gradient explosion or diminishing problems when the number of inner loop iterations and model depth increase~\cite{antoniou2018how}, and both are inevitable for training summarization models.
Third, MAML requires diverse source tasks to increase the generalizability on novel target tasks. However, how to build such meta-dataset on existing corpora for summarization tasks still remains unknown.

To solve these challenges, we propose a simple yet effective method, named \textbf{Meta-Transfer Learning for low-resource ABStractive summarization (MTL-ABS)}\footnote{Code is available at https://github.com/YiSyuanChen/MTL-ABS}. Specifically, to address the first and second challenges, we propose using a limited number of parameters and layers between layers of a pre-trained network to perform meta learning. An alternative approach is to stack or replace several new layers on the top of the pre-trained model and only meta-learn these layers to control the model complexity. However, without re-training the full model, the performance may significantly drop due to the introduction of consecutive randomly initialized layers. Moreover, it is difficult to recover the performance by using limited target labeled examples for re-training. In addition to the limited number of parameters and layers between layers, to better leverage the pre-trained model, we add skip-connections to the new layers. With small initialization values, it alleviates the interference at the beginning of training. Also, with a limited number of new layers, the complexity of the framework can be reduced as a compact model with skip-connections, which can prevent gradient problems during meta learning. 

For the third challenge, most existing methods use inherent labels in a single dataset to handle this problem. For instance, in the few-shot image classification \cite{Sun_2019_CVPR}, the tasks are defined with different combinations of class labels. However, this strategy is not applicable for abstractive summarization since there is no specific label to characterize the property for article-summary pairs. One possible solution is to randomly sample data from a single corpus such that the inherent data variance provides task diversity. Take this idea further, we consider exploring multiple corpora to increase the diversity of different tasks. Since the data from different corpus could have distributional differences, an inappropriate choice of source corpora may instead lead to a negative transfer problem and deteriorate the performance. Thus, we investigate this problem by analyzing the performance of different corpora choices. Specifically, we study several similarity criteria and show that some general rules can help avoid inappropriate choices, which is crucial in developing meta learning for NLP tasks. The contributions are summarized as follows.

\begin{itemize}

\item Beyond conventional methods that only utilize a single large corpus, we further leverage multiple corpora to improve the performance. To the best of our knowledge, this is the first work to explore this opportunity with meta-learning methods for abstractive summarization.

\item We propose a simple yet effective method, named MTL-ABS, to tackle the low-resource abstractive summarization problem. With the proposed framework, we make successful cooperation of transfer learning and meta learning. Besides, we investigate the problem of choosing source corpora for meta-datasets, which is a significant but not well-studied problem in meta learning for NLP. We provide some general criteria to mitigate the negative effect from inappropriate choices of source corpora.

\item Experimental results show that MTL-ABS achieves the state-of-the-art on 6 corpora in low-resource abstractive summarization, with only 0.7\% of trainable parameters compared to previous work.

\end{itemize}

\section{Related Works}
Transfer learning has been widely adopted to tackle applications with limited data. For NLP tasks, word representations are usually pre-trained by self-supervised learning with unlabeled data, and then used as strong priors for downstream tasks. Among the pre-training methods, language modeling (LM) \cite{devlin-etal-2019-bert,radford2019language,radford2018improving} has achieved great success. To transfer with pre-trained models for downstream tasks, it is common to add some task-specific layers on top and fine-tune the full model. However, this strategy is often inefficient with regard to the parameter usage, and full re-training may be required when encountering new tasks.  Thus, \citet{pmlr-v97-houlsby19a} propose a compact adapter module to transfer from the BERT model for natural language understanding tasks. Each BERT layer is inserted with few adapter modules, and only the adapter modules are set to be learnable. \citet{pmlr-v97-stickland19a} similarly transfer from BERT with Projected Attention Layers (PALs) for multi-task natural language understanding, which is a multi-head attention layer residually-connected to the base model.

For abstractive summarization, \citet{liu-lapata-2019-text} propose a Transformer based encoder-decoder framework. The training process includes two-level pre-training. The encoder is first pre-trained with unlabelled data as a language model, then fine-tuned to perform extractive summarization tasks. Finally, the decoder is added to learn for abstractive summarization tasks. \citet{pmlr-v119-zhang20ae} further propose to pre-train the decoder in a self-supervised way with the Gap Sentence Generation (GSG) task. It selects and masks important sentences according to the ROUGE scores of sentences and the rest of the article, and the objective for the decoder is to reconstruct the masked sentences. 

While abstractive summarization performance is improved with various transfer learning techniques, there is much less literature regarding the low-resource setting. \citet{radford2019language} propose a Transformer based language model trained on a massive-scale dataset consisting of millions of webpages, and the abstractive summaries are produced based on the pre-trained generative ability under the zero-shot setting. \citet{DBLP:journals/corr/abs-1905-08836} propose a Transformer based decoder-only language model with newly collected data from Wikipedia. \citet{pmlr-v119-zhang20ae} report relatively outstanding performance on various datasets in low-resource settings, with the specially designed pre-training objective for abstractive summarization. However, these works only use a single large corpus for training, which can suffer severe domain-shifting problems for some target corpus. Besides, the large model size could cause an overfitting problem. On the contrary, our framework uses limited trainable parameters with multiple corpora chosen according to the proposed criteria to mitigate the above problems.

\section{Methodologies}
In this work, we define the low-resource abstractive summarization problem as follows:
\begin{definition}{\textbf{Low-Resource Abstractive Summarization}}
is a task that requires a model to learn from experience $E$, which consists of direct experience $E_d$ containing limited monolingual article-summary pairs and indirect experience $E_i$, to improve the performance in abstractive summarization measured by the evaluation metric $P$.  
\end{definition}
The direct experience $E_d$ refers to the training examples of target corpus, and the indirect experience $E_i$ could be other available resources such as pre-trained models or related corpora. For measurement $P$, we use ROUGE \cite{lin2004rouge} for evaluation. In this work, we consider a challenging scenario that the quantity of labeled training examples for the target corpus is limited under 100, which matches the magnitude of the evaluation-only Document Understanding Conference (DUC) corpus. In the following, we first introduce the proposed summarization framework, and then elaborate on the meta-transfer learning process and the construction of meta-dataset.

\begin{figure}[t]
\centering
\includegraphics[width=0.48\textwidth]{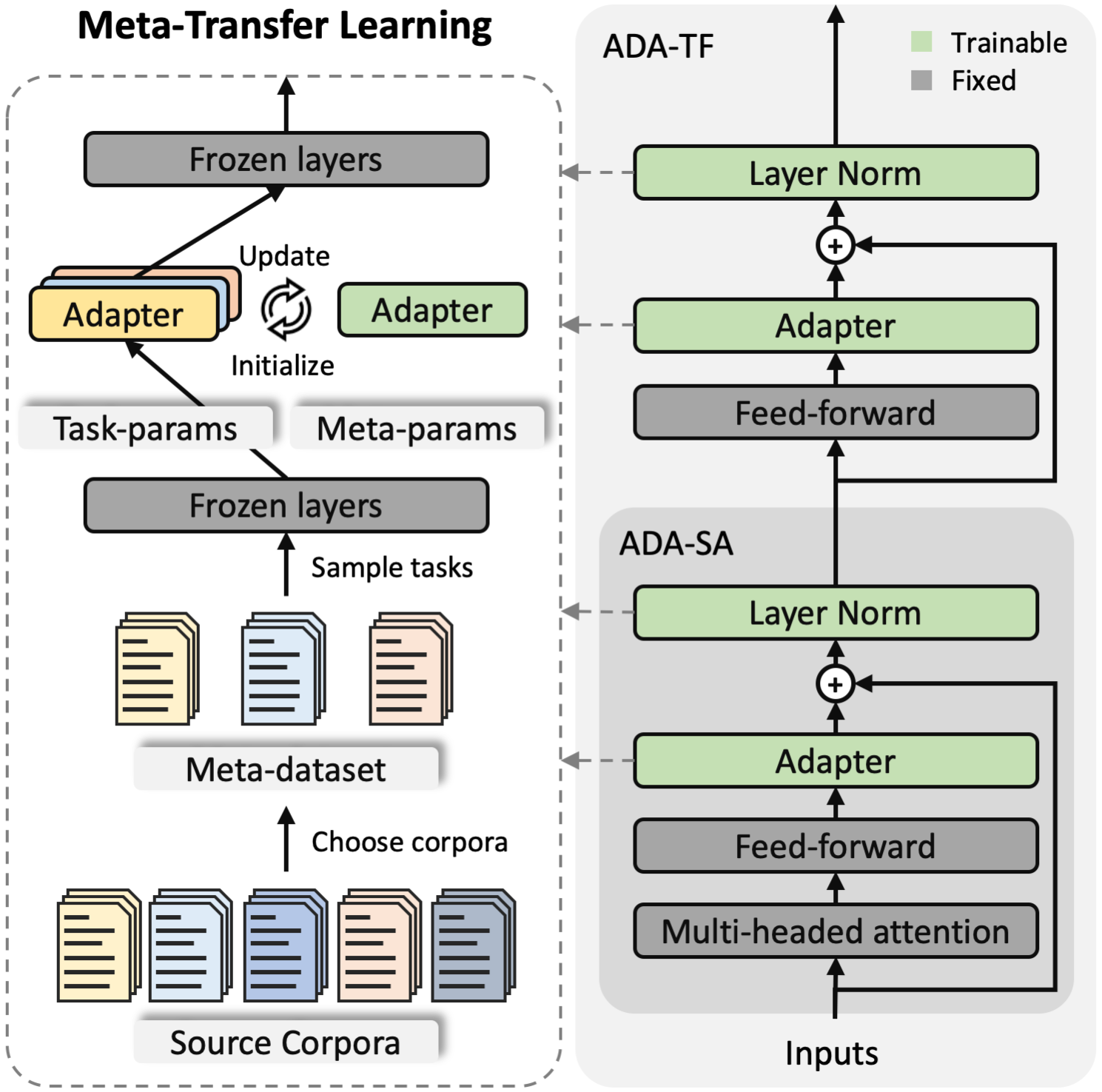} 
\caption{Proposed summarization framework with meta-transfer learning. The adapter modules are inserted into both encoder and decoder after every feed-forward layer. During meta-transfer learning, only the adapters and layer normalization layers are learnable. For simplicity, the learning illustration of layer normalization layers is omitted.}
\label{fg:framework}
\end{figure}

\subsection{Summarization Framework}

\subsubsection{Base Model}
We choose the state-of-the-art Transformer based encoder-decoder model \cite{liu-lapata-2019-text} as the base summarization model. Special token [CLS] is added at the beginning of each sentence to aggregate information, while token [SEP] is appended after each sentence as boundary. The self-attention (SA) layer mainly consists of three sub-layers, which are the Multi-Headed Attention (MHA) layer, Feed-Forward (FF) layer, and Layer Normalization (LN) layer. The self-attention layer can be expressed as:
\begin{equation}
\text{SA}(h)=\text{LN}(\text{FF}(\text{MHA}(h)) + h),
\end{equation}
where $h$ represents the intermediate hidden representation. The transformer (TF) layer is stacked with self-attention layers and can thus be expressed as:
\begin{equation}
\text{TF}(h)= \text{LN}(\text{FF}(\text{SA}_{1:l}(h))),
\end{equation}
where $l$ indicates the number of self-attention layers. We set $l=1$ for the encoder and $l=2$ for the decoder. 
The encoder of the base model is initialized with BERT \cite{devlin-etal-2019-bert}, which is trained on the general domain. Before the meta-transfer learning, we fine-tune the encoder with an extractive objective on the chosen pre-training corpus, as previous works suggest \cite{liu-lapata-2019-text,li-etal-2018-improving,gehrmann-etal-2018-bottom}, to improve the abstractive performance.

\subsubsection{Adapters}
To prevent overfitting and gradient instability from applying MAML on the large pre-trained model, we propose restricting the number of meta-trainable parameters and layers. This is practically achieved with adapter modules. The adapter module is a bottlenecked feed-forward network consisting of a down-project layer $f_{\theta_d}$ and an up-project layer $f_{\theta_u}$. A skip-connection from input to output is established, which prevents the noised initialization from interference with the training initially. The adapter (ADA) can be expressed as: 
\begin{equation}
    \text{ADA}(h) = f_{\theta_u}(\text{ReLU}(f_{\theta_d}(h))) + h.
\end{equation}
We insert adapters into each layer of the encoder and decoder to leverage pre-trained knowledge while performing meta learning. Specifically, the adapter is added after every feed-forward layer in the transformer layer. Thus, the adapted transformer (ADA-TF) layers with adapted self-attention (ADA-SA) layers can be expressed as:
\begin{equation}
\text{ADA-SA}(h)=\text{LN}(\text{ADA}(\text{FF}(\text{MHA}(h)) + h)),
\end{equation}
\begin{equation}
\text{ADA-TF}(h)=\text{LN}(\text{ADA}(\text{FF}(\text{ADA-SA}_{1:l}(h)))).
\end{equation}
The illustration of the proposed summarization framework is shown in Figure~\ref{fg:framework}.

\subsection{Meta-Transfer Learning for Summarization}
Equipped with the adapter-enhanced summarization model, our goal is to perform the meta-transfer learning for fast adaption on new corpora. Consider a pre-training corpus $C^{pre}$, a set of source corpora $\{ C^{src}_j \}$ and a target corpus $C^{tgt}$, we aim to leverage both $C^{pre}$ and $\{ C^{src}_j \}$ to improve the performance on $C^{tgt}$, which only contains limited number of labeled examples. For abstractive summarization, a training example consists of input, prediction and ground-truth sequences. We denote them by $\mathcal{X}=[X_1,...,X_{N_x}]$, $\mathcal{Y}=[Y_1,...,Y_{N_y}]$, and $\widehat{\mathcal{Y}}=[\widehat{Y}_1,...,\widehat{Y}_{N_y}]$, respectively.

Our framework comprises the base summarization model $\theta$ and the adapter modules $\psi$. The two parts are learned in decoupled scheme. For the base model, given an input article $\mathbf{x}=[x_1,...,x_{N_x}] \in \mathcal{X}$, the model produces a particular prediction sequence $\mathbf{y}_{<t}=[y_1,...,y_{t-1}]$ at time $t$, and the probability to generate token $y_t$ is defined as:
\begin{multline}
    p(y_t|\mathbf{y}_{<t},\mathbf{x},\theta) \doteq Pr(Y_t=y_t|Y_{t-1}=y_{t-1},..., \\ Y_{1}=y_1,X_{N_x}=x_{N_x},...X_1=x_1,\theta).
\end{multline}
With the ground-truth summary $[\widehat{y}_1,...,\widehat{y}_{N_y}] \in \mathcal{\widehat{Y}}$, we optimize the model to minimize the negative log-likelihood (NLL) as:
\begin{align}
    -logp(\mathbf{y}|\mathbf{x},\theta)  = -\sum^{N_y}_{t=1}logp(\widehat{y}_t|\mathbf{y}_{<t},\mathbf{x},\theta),
\end{align}
\begin{equation}
    -logp(C^{pre}|\theta) = -\sum_{(\mathbf{x},\mathbf{y}) \in C^{pre}}logp(\mathbf{y}|\mathbf{x},\theta).
\end{equation}

After the training of the base model, we insert adapter modules into the framework. To meta-learn the adapter modules, we sample examples without replacement from a set of source corpora $\{ C^{src}_j \}$ to create a collection of tasks $\{\mathcal{T}_i\}_{i=1}^M$ as meta-dataset $\mathcal{M}$. The source corpora are chosen according to the proposed criteria, which will be introduced in the next section. For each task $\mathcal{T}_i$, it contains a task-training set $D^{tr}_i=\{ (\mathbf{x}^k_i,\mathbf{y}^k_i) \}^K_{k=1}$ and a task-testing set $D^{te}_i=\{ (\mathbf{x}^{*k}_i, \mathbf{y}^{*k}_i) \}^K_{k=1}$. The number of tasks from different corpus is further balanced to avoid domain bias. The meta learning process includes the two-level optimization. At each meta-step, we consider a batch of tasks $\mathcal{B}=\{\mathcal{T}_i | i \in B\}$. In the inner loop optimization, the base-learner $\phi$ is initialized with $\psi$ and minimizes the following objective with task-training set $D^{tr}_i$: 
\begin{equation}
    -\log p(D^{tr}_i|\phi,\psi) = -\sum_{(\mathbf{x},\mathbf{y}) \in D^{tr}_i}\log p(\mathbf{y}|\mathbf{x},\phi,\psi).
\end{equation}
Through the optimization, the task-parameters $\phi$ will be adapted to the specific task $i$ as $\phi_i$. Assume there is only one update step, it can be expressed as: 
\begin{equation}
    \phi_i \leftarrow \psi + \beta\nabla_\phi \log p(D^{tr}_i|\phi,\psi).
\end{equation}
For the outer loop optimization, meta-learner $\psi$ in another way minimizes the testing loss after adaptation with all task-testing set $D^{te}_i$ as follows:
\begin{equation}
    -\log p(\mathcal{B}|\psi)=-\sum_{i \in B} \log p(D^{te}_i|D^{tr}_i,\phi_i,\psi).
\end{equation}
$\psi$ is then updated by:
\begin{equation}
    \psi \leftarrow \psi + \alpha \nabla_\psi \log p(\mathcal{B}|\psi).
\end{equation}
After each meta-step, the meta-parameters $\psi$ will possess more generalization knowledge from different tasks.

The base-learner and meta-learner can be optimized with SGD or other momentum-based optimizers such as Adam \cite{DBLP:journals/corr/KingmaB14}. Practically, we observe that tasks from different corpora could have distributional differences, and sharing the same momentum-based optimizer in the inner loop could lead to slow convergence. Therefore, we use separated optimizers for tasks from different corpora and accumulate the momentum statistics within the same corpus to accelerate the training. The left part of Figure~\ref{fg:framework} illustrates the process of proposed meta-transfer learning.  

\subsection{Meta-Datasets}
To address the third challenge, it requires creating diverse source tasks to increase the generalizability on novel target tasks. In applications such as classification \cite{Sun_2019_CVPR,obamuyide-vlachos-2019-model}, the tasks for MAML can be easily defined with class labels from a single corpus. However, it is not applicable to abstractive summarization. Instead of randomly sampling data from a single corpus to construct tasks, we propose to leverage multiple corpora. Specifically, since the tasks are defined as data from different corpora, the problem thus becomes how to choose source corpora. The intuitive idea is to choose diverse source corpora. Meanwhile, each source corpus should possess as many similar identities to the target corpus as possible. A suitable similarity criterion should show a monotonous performance change with corpora chosen along the similarity ranking. Based on this idea, we consider the following hypotheses that may help in source corpora choice:

\begin{itemize}

    \item \textbf{Semantics.} Source corpus that similar to the target corpus in semantics may provide better knowledge to comprehend articles and identify salient contents for abstractive generations. We quantify this property with document embedding similarity.
    
    \item \textbf{Word Overlapping.} Source corpus with high word overlap to target corpus may provide more primary knowledge to use precise words. We quantify this property with cosine similarity.
    
    \item \textbf{Coverage.} Source corpus that covers as many used words in the target corpus may provide more transferable knowledge. We quantify this property with ROUGE recall.
    
    \item \textbf{Information Density.} The information density is defined as the number of word overlap divided with the number of words in the source corpus. High information density indicates that the source corpus contains a large proportion of transferable knowledge. We quantify this property with ROUGE precision.
    
    \item \textbf{Length.} The length of an article can reflect the amount of information. Source corpus with a similar average length to the target corpus can help reduce the distributional differences. We quantify this property with an absolute difference of token length between articles.

\end{itemize}

Given a target corpus, we rank the source corpora and create meta-datasets in different ranking sections for each criterion. Next, we conduct experiments to investigate the relation between performance and source corpora choice. The analysis and implementation are detailed in the experiment section. According to the results, we have the following observations. First, the statistical similarity is more important than the semantic similarity.
Second, corpora from the same domain may not always provide better knowledge. 
Third, rather than choosing source corpora that can cover all used words in the target corpus, source corpora should have a high information density. Fourth, source corpora with similar average article length are better. With the above observations, we use the average ranking of the following criteria: 1) cosine similarity, 2) ROUGE precision, and 3) article length to choose our source corpora for meta-dataset.

\section{Experiments}

\subsection{Implementation Details} 
All of our experiments are conducted on a single NVIDIA Tesla V100 32GB GPU with PyTorch. The self-attention layer we used has 768 hidden neurons with 8 heads, and the feed-forward layer contains 3072 hidden neurons. The encoder consists of 12 transformer layers with a dropout rate of 0.1, and the decoder has 6 transformer layers with a dropout rate of 0.2. For adapter modules, the hidden size is 64. The vocabulary size is set to 30K. For meta-training, unless otherwise specified, a meta-batch includes 3 tasks, and the batch size for each task is 4. The base-learner and meta-learner are both optimized with Adam \cite{DBLP:journals/corr/KingmaB14} optimizer, and the learning rate is set to 0.0002. The inner gradient step is 4, and the whole model is trained with 6K meta-steps. For meta-validation, we use a corpus excluded from source tasks and target task, and the performance is calculated as an average of 600 batches.

\renewcommand\arraystretch{1.0}
\begin{table*}[t]
\centering\fontsize{9}{11}\selectfont
\begin{tabular}{p{0.1\textwidth}>{\centering}p{0.08\textwidth}>{\centering}p{0.15\textwidth}>{\centering}p{0.15\textwidth}>{\centering}p{0.15\textwidth}>{\centering\arraybackslash}p{0.17\textwidth}}

\toprule
\multirow{2}{*}{Dataset} & \multirow{2}{*}{\makecell{Labeled \\ examples}} & \citet{pmlr-v119-zhang20ae} & TL-ABS & MTL-ABS & Improving ratio  \\
& & $R_1$ / $R_2$ / $R_L$ & $R_1$ / $R_2$ / $R_L$ & $R_1$ / $R_2$ / $R_L$ & $R_1$ / $R_2$ / $R_L$ \\
 
\midrule
\multirow{2}{*}{AESLC} & 10 & 11.97/4.91/10.84 & 17.32/8.07/16.82 & \textbf{21.27/10.79/20.85} & +78\% / +120\% / +92\% \\
& 100 & 16.05/7.20/15.32 & 21.59/10.48/20.92 & \textbf{23.88/12.06/23.18} & +49\% / +68\% / +51\% \\
\midrule
\multirow{2}{*}{BillSum} & 10 & 40.48/18.49/\textbf{27.27} & 40.06/17.66/26.62 & \textbf{41.22/18.61}/26.33 & +2\% / +0.6\% / -3\% \\
& 100 & 44.78/\textbf{26.40/34.40} & 44.12/21.64/29.14 & \textbf{45.29}/22.74/29.56 & +1\% / -14\% / -14\% \\
\midrule
\multirow{2}{*}{Gigaword} & 10 & 25.32/8.88/22.55 & 26.67/10.04/24.42 & \textbf{28.98/11.86/26.74} & +14\% / +34\% / +19\% \\
& 100 & 29.71/12.44/27.30 & 29.54/12.22/27.22 & \textbf{30.03/12.70/27.71}  & +1\% / +2\% / +2\% \\
\midrule
\multirow{2}{*}{Multi-News} & 10 & \textbf{39.79}/12.56/\textbf{20.06} & 37.8/11.48/20.92 & 38.88/\textbf{12.78}/19.88 & -2\% / +2\% / -1\% \\
& 100 & \textbf{41.04/13.88/21.52} & 38.82/13.03/20.62 & 39.64/13.64/20.45 & -3\% / -2\% / -5\% \\
\midrule
\multirow{2}{*}{NEWSROOM} & 10 & 29.24/17.78/24.98 & 30.43/15.87/25.93 & \textbf{37.15/25.40/33.78} & +27\% / +43\% / +35\% \\
& 100 & 33.63/21.81/29.64 & 29.96/17.34/26.27 & \textbf{41.86/30.10/38.26} & +24\% / +38\% / +29\% \\
\midrule
\multirow{2}{*}{Reddit-TIFU} & 10 & 15.36/2.91/10.76 & 16.68/5.16/15.63 & \textbf{18.03/6.41/17.10} & +17\% / +120\% / +59\% \\
& 100 & 16.64/4.09/12.92 & 18.06/6.75/17.29 & \textbf{20.14/7.71/19.38} & +21\% / +89\% / +50\% \\
\midrule
\multirow{2}{*}{arXiv} & 10 & 31.38/8.16/17.97 & 34.59/8.37/19.16 & \textbf{35.81/10.26/20.51} & +14\% / +26\% / +14\% \\
& 100 & 33.06/9.66/20.11 & 36.61/9.83/20.00 & \textbf{37.58/10.90/20.23} & +14\% / +13\% / +0.6\% \\
\midrule
\multirow{2}{*}{PubMed} & 10 & 33.31/\textbf{10.58}/\textbf{20.05} & 32.96/9.10/20.20 & \textbf{34.08}/10.05/18.66 & +2\% / -5\% / -7\% \\
& 100 & 34.05/\textbf{12.75}/\textbf{21.12} & 35.11/11.06/20.14 & \textbf{35.19}/11.44/19.89 & +3\% / -10\% / -6\% \\
\midrule
\multirow{2}{*}{WikiHow} & 10 & 23.95/6.54/15.33 & 28.09/7.69/19.67 & \textbf{28.34/8.16/19.72} & +18\% / +25\% / +29\% \\
& 100 &25.24/7.52/17.79 & 29.48/8.38/20.03 & \textbf{31.00/9.68/21.50} & +23\% / +29\% / +21\% \\
 
\bottomrule
\end{tabular}
\caption{Low-resource performance of MTL-ABS on all target corpora compared with PEGASUS. We pre-process all corpora according to PEGASUS for comparison. Best ROUGE numbers on each corpus are bolded, and improving ratios for PEGASUS to MTL-ABS are also shown. The number of trainable parameters for MTL-ABS and TL-ABS is 4.23M, and for PEGASUS is 568M. The vocabulary size is 30K and 96K for MTL-ABS and PEGASUS, respectively.}\smallskip
\label{tb:low_resource_performance}

\end{table*}

\subsection{Corpora} 
We use following corpora to verify proposed methods. Note that corpus chosen as target task will not be included in the source corpora.
\begin{itemize}
\item{\textbf{AESLC}} \cite{zhang-tetreault-2019-email} is a collection of 18K email messages and corresponding subject lines of employees in the Enron Corporation. 


\item{\textbf{BillSum}} \cite{kornilova-eidelman-2019-billsum} contains 22K US bills with human-written summaries from the 103rd-115th (1993-2018) sessions of Congress, and 1K California bills from the 2015-2016 session.

\item{\textbf{CNN/DailyMail}} \cite{NIPS2015_afdec700} contains 93k and 220k news articles with multiple-sentence summaries from CNN and Daily Mail newspapers, respectively. We use this corpus to pre-train our summarization framework. 


\item{\textbf{Gigaword}} \cite{rush-etal-2015-neural} is a headline-generation corpus that contains 4M news articles and headlines sourced from various services.

\item{\textbf{Multi-News}} \cite{fabbri-etal-2019-multi} is a large-scale dataset for multi-document summarization, which contains 56K articles from diverse news sources accompanied by human-written summaries. 

\item{\textbf{NEWSROOM}} \cite{grusky-etal-2018-newsroom} contains 1.3 million articles and summaries written by authors and editors in newsroom. The summaries are written in various strategies, including extraction and abstraction.

\item{\textbf{Webis-TLDR-17}} \cite{volske-etal-2017-tl} contains 4M posts from Reddit with author-provided "TL;DR” as summaries. This corpus is only used as a source task since \citet{pmlr-v119-zhang20ae} do not report the results. 

\item{\textbf{Reddit-TIFU}} \cite{kim-etal-2019-abstractive} contains 123K posts and corresponding summaries especially from TIFU subreddit, which are more casual and conversational.

\item{\textbf{arXiv, PubMed}} \cite{cohan-etal-2018-discourse} are long-document summarization datasets collected from scientific repositories. It contains 215K and 133K articles and corresponding abstracts from arXiv and PubMed, respectively.

\item{\textbf{WikiHow}} \cite{DBLP:journals/corr/abs-1810-09305} contains 230K articles and summaries written by different authors from WikiHow.


\end{itemize}

\subsection{Low-Resource Performance} 
We compare the low-resource performance of the proposed Meta-Transfer Learning for low-resource ABStractive summarization (MTL-ABS) with two baselines. One is a naive transfer learning method applied to the proposed summarization framework, denoted as TL-ABS. In other words, TL-ABS only finetunes the adapter modules with labeled target examples. The other baseline is PEGASUS \cite{pmlr-v119-zhang20ae}, which is a large pre-trained encoder-decoder framework. The encoder's learning objective is the conventional Mask Language Model (MLM), while the decoder's objective is Gap Sentences Generation (GSG) that is specifically designed for abstractive summarization. For each target corpus, we use the rest of the corpora as candidates to build meta-datasets, and the number of examples in each source corpus is limited under 40K for balance. The combination of source corpora in meta-dataset is decided according to the proposed criteria. For adaptation, we finetune the meta-learned model with 10 or 100 labeled examples on the target corpus. ROUGE is used as the evaluation metric. 

Table~\ref{tb:low_resource_performance} compares MTL-ABS with TL-ABS and PEGASUS on different datasets in terms of ROUGE score. The results manifest that MTL-ABS outperforms PEGASUS on 6 out of 9 corpora and achieves compatible results on the other three corpora. On AESLC and Reddit-TIFU, MTL-ABS achieves ROUGE2-F of 10.79 and 6.41, which improves the performance of PEGASUS for 120\%. For NEWSROOM and WikiHow, the performance improvements are over 18\% for all evaluation metrics. MTL-ABS performs relatively weak on lengthy summarization corpora such as BillSum, Multi-News, and PubMed. To investigate this problem, we follow previous work \cite{grusky-etal-2018-newsroom,pmlr-v119-zhang20ae} to calculate the extractive fragment coverage and density for all source corpora. The results show that BillSum, Multi-News, and Pubmed are the top-three in average density, indicating higher extractive property. While the architecture and learning process of MTL-ABS mainly focus on the abstractive generation, we consider that this problem can be further solved with techniques such as copy mechanism \cite{see-etal-2017-get}, which would serve as our future work to improve the performance on these corpora. Comparing the performance of MTL-ABS to TL-ABS, the meta-learned initialization works especially well on the scenario of 10 labeled examples. This suggests that our method can be suitable for extreme cases that suffer from severe data scarcity problems. It is also worth noting that MTL-ABS uses only 4.23M training parameters, while PEGASUS uses 568M, demonstrating the parameter efficiency of the proposed framework.

\subsection{Preventing Overfitting Problem} 
The task size of MAML is often set to be small for computation efficiency. However, this could prevent MAML from utilizing large models due to the overfitting problem. MTL-ABS alleviates this problem with the restriction of learnable parameters. Table~\ref{tb:overfitting} shows that full model fine-tuning (TL-FULL) could easily overfit and lead to worse generalizability. The performance is even inferior to naive transfer learning methods with the proposed framework (TL-ABS).

\renewcommand\arraystretch{1.0}
\begin{table}[t]
\centering\fontsize{9}{11}\selectfont
\begin{tabular}{lccc}

\toprule
\multirow{2}{*}{Dataset} & TL-FULL & MTL-ABS \\
& $R_1$ / $R_2$ / $R_L$ & $R_1$ / $R_2$ / $R_L$ \\
\midrule
AESLC & 15.42/7.20/15.12  & \textbf{21.27/10.79/20.85}  \\
\midrule
Gigaword & 25.15/8.89/23.07 & \textbf{28.98/11.86/26.74} \\
\midrule
Reddit-TIFU & 13.90/4.00/13.32 & \textbf{18.03/6.41/17.10} \\

\bottomrule
\end{tabular}
\caption{ Performance of MTL-ABS compared with fine-tuning on full model (TL-FULL) with 10 labeled examples. }\smallskip
\label{tb:overfitting}
\end{table}

\subsection{Preventing Gradient Problem in Meta Learning}
\begin{figure}[t]
\centering
\includegraphics[width=0.43\textwidth]{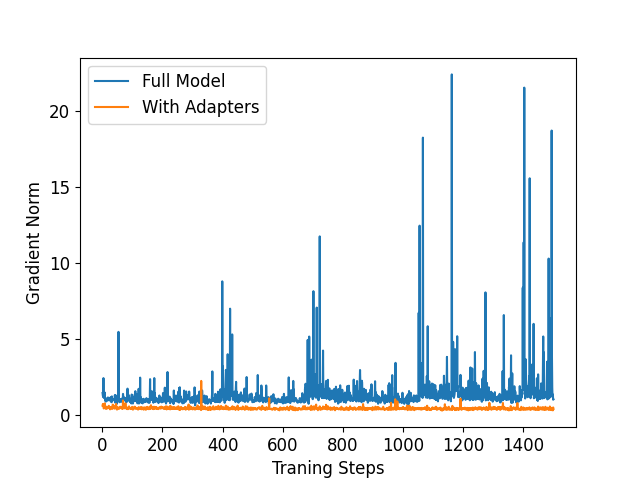} 
\caption{Dynamics of gradient norm along the meta-transfer learning process with 1) full model is trainable, and 2) only adapter modules are trainable. }
\label{fg:grad_norm}
\end{figure}

As pointed out by previous work \cite{antoniou2018how}, deep models with many inner loop iterations can cause gradient explosion and diminishing in meta learning. MTL-ABS alleviates this problem by utilizing adapter modules with skip-connections. Figure~\ref{fg:grad_norm} shows the gradient norm dynamics for the proposed framework and the model that meta-learns on all parameters. The results show that the gradient norm is unstable in full model meta-transfer learning, and the training fails due to numerical problems after 1500 meta-steps.

\subsection{Impact on the Choice of Source Corpora}
\label{subsec:src_corpora_choice}
To study the relation of performance and source corpora choice, we choose AESLC as our target corpus, and other corpora are the candidates to build meta-dataset. The source corpora are first ranked with the following five criteria: 1) document embedding similarity, 2) cosine similarity, 3) length similarity, 4) ROUGE-2 recall, and 5) ROUGE-2 precision. We use the encoder of our pre-trained summarization framework to extract document embeddings, and the similarity of embeddings is calculated as a normalized inner product. The cosine similarity is computed as $n(S_A \cap S_B)/\sqrt{n(S_A) * n(S_B)}$, where $S_A$ and $S_B$ are word sets of compared articles, and $n(S)$ is the number of words in the set. The ROUGE-2 recall and precision are calculated as (common 2-grams/target 2-grams) and (common 2-grams/source 2-grams) respectively. For length similarity, we compute $abs(L_A-L_B)$, where $L_A$ and $L_B$ are the token lengths of compared articles. The similarity results are shown in Table~\ref{tb:sim}. Next, we create five meta-datasets for each criterion with different sections along ranking, i.e., [1-3], [3-5], [4-6], [5-7], and [7-9], to investigate the performance.

\renewcommand\arraystretch{1.0}
\begin{table}[t]
\centering\fontsize{9}{11}\selectfont
\begin{tabular}{p{0.1\textwidth}>{\centering\arraybackslash}p{0.25\textwidth}}

\toprule
Criteria &  Similarity (High $\rightarrow$ Low) \\ 
\midrule
Embedding & \textbf{B},\textbf{M},\textbf{N},\textbf{P},\textbf{W},\textbf{X},\textbf{L},\textbf{G},\textbf{T} \\ 
\midrule
Cosine & \textbf{L},\textbf{T},\textbf{W},\textbf{N},\textbf{M},\textbf{X},\textbf{P},\textbf{B},\textbf{G} \\ 
\midrule
Length & \textbf{G},\textbf{L},\textbf{T},\textbf{W},\textbf{N},\textbf{B},\textbf{X},\textbf{P},\textbf{M} \\ 
\midrule
ROUGE-2-R & \textbf{M},\textbf{P},\textbf{X},\textbf{W},\textbf{N},\textbf{T},\textbf{B},\textbf{L},\textbf{G} \\ 
\midrule
ROUGE-2-P & \textbf{W},\textbf{L},\textbf{T},\textbf{N},\textbf{X},\textbf{G},\textbf{M},\textbf{P},\textbf{B} \\ 

\bottomrule
\end{tabular}
\caption{Similarity rankings for target corpora AESLC with source corpora BillSum, Gigaword, Multi-News, NEWSROOM, TLDR-17, Reddit-TIFU, arXiv, PubMed and WikiHow. }\smallskip
\label{tb:sim}
\end{table}

\begin{figure}[t]
\centering
\includegraphics[width=0.47\textwidth]{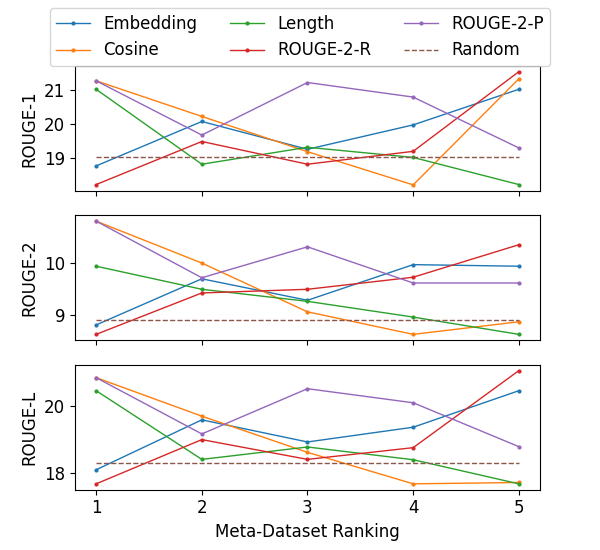} 
\caption{Performance comparison of different similarity criteria including 1) document embedding similarity, 2) cosine similarity and 3) length similarity, 4) ROUGE-2 recall, and 5) ROUGE-2 precision for target corpus AESLC with 10 labeled examples. Best viewed in colors.}
\label{fg:ranking}
\end{figure}

Figure~\ref{fg:ranking} shows that the performance varies a lot with different similarity criteria. Some of them are even worse than random choice. From the comparison of embedding, cosine, and length similarities, it shows that the performance is more correlated to the word overlap and article length. For the word overlap, the results also show that ROUGE-2 precision is a better criterion than ROUGE-2 recall, which means that the source corpus should have higher information density (word overlap in source and target/words in source) rather than covering all words in target corpus. In other words, it is better that the source corpus does not contain too much out-of-distribution information. In addition to the content of article, the length can also be an influential factor. Interestingly, the Gigaword corpus is ranked last in cosine similarity but top in length similarity, while including Gigaword can actually improve the performance. In this case, word intersection shows less indicative when the source corpus' length is similar to the target corpus. 

To further verify the above observations, we also use Gigaword as a target corpus to conduct experiments with the same settings. The results show that Gigaword is most similar in document embedding with Multi-News and NEWSROOM, which meets expectations since these corpora are in the news domain. However, the performance results show that Gigaword is better benefited from WikiHow and AESLC with a ROUGE-2 of 11.77, while it is 10.23 for Multi-News and NEWSROOM. It indicates that source corpora from the same domain may not always give better transfer knowledge. In conclusion, we report the best performance in Table~\ref{tb:low_resource_performance} with top-3 corpora using the average ranking of the following criteria: 1) cosine similarity, 2) ROUGE-2 precision, and 3) article length.

\section{Conclusion}
In this work, we propose a simple yet effective meta-transfer learning method for low-resource abstractive summarization. We effectively combine the transfer learning and meta learning by using adapter modules as the bridge. Moreover, we also investigate and provide general criteria for source corpora choice, which is a field that has not been studied in meta learning for NLP. Experimental results demonstrate that the proposed method outperforms the state-of-the-art on 6 diverse corpora. In the future, we plan to explore methods for better leveraging distant source corpus, which is important since there is no guarantee for the availability of similar source corpora. Second, we plan to extend this framework to modulate the pre-trained parameters for better adaption on novel tasks.

\section{Acknowledgements}
\label{sec:ack} 
We are grateful to the National Center for High-performance Computing for computer time and facilities. This work was supported in part by the Ministry of Science and Technology of Taiwan under Grants MOST-109-2221-E-001-015, MOST-109-2218-E-009-016, and MOST-109-2221-E-009-114-MY3.

\bibliography{main}

\end{document}